\documentclass{article}
\usepackage{spconf,amsmath,graphicx}

\usepackage{framed,multirow}
\usepackage{amssymb}
\usepackage{latexsym}
\usepackage{url}
\usepackage{xcolor}
\definecolor{newcolor}{rgb}{.8,.349,.1}
\usepackage{times}
\usepackage{epsfig}
\usepackage{graphicx}
\usepackage{subcaption}
\usepackage{amsmath}
\usepackage{multicol}
\usepackage{multirow}
\usepackage{ textcomp }
\usepackage{comment}
\usepackage{xspace}

\usepackage[breaklinks=true,bookmarks=false]{hyperref}

\newcommand{\etal}{\textit{et al.}\xspace}
\newcommand{\ie}{\textit{i.e.}\xspace}

\newcommand{\slfrac}[2]{\left.#1\middle/#2\right.}

\title{Exploring Attention Mechanisms in Integration of Multi-Modal Information for Sign Language Recognition and Translation}
\name{Zaber Ibn Abdul Hakim, Rasman Mubtasim Swargo, Muhammad Abdullah Adnan}
\address{Bangladesh University of Engineering and Technology (BUET), Dhaka, Bangladesh}

\begin{document}

\maketitle

\begin{abstract}
Understanding intricate and fast-paced movements of body parts is essential for the recognition and translation of sign language. The inclusion of additional information intended to identify and locate the moving body parts has been an interesting research topic recently. However, previous works on using multi-modal information raise concerns such as sub-optimal multi-modal feature merging method, or the model itself being too computationally heavy. In our work, we have addressed such issues and used a plugin module based on cross-attention to properly attend to each modality with another. Moreover, we utilized 2-stage training to remove the dependency of separate feature extractors for additional modalities in an end-to-end approach, which reduces the concern about computational complexity. Besides, our additional cross-attention plugin module is very lightweight which doesn't add significant computational overhead on top of the original baseline. We have evaluated the performance of our approaches on the RWTH-PHOENIX-2014 dataset for sign language recognition and the RWTH-PHOENIX-2014T dataset for the sign language translation task. Our approach reduced the WER by 0.9 on the recognition task and increased the BLEU-4 scores by 0.8 on the translation task.
\end{abstract}
\section{Introduction}
\label{sec:intro}

1.5 billion people on this earth have some form of hearing impairment, and statistically, eight out of 100,000 people are born mute \cite{SLSchool}, and the most important medium to communicate with these deaf-mute people is through Sign Language (SL). Sign languages contain both visual manual (different hand shapes and movements) and visual non-manual (head movement, facial expressions) features. SLs are manifested by complex gestures and fast movements of body parts which makes it hard for mass people to get used to. In such a scenario, it has become imperative to devise an automatic approach to understand such movements and translate sign languages into human-readable format. For this reason, automatic sign language recognition and translation have gained widespread attention not only for academic purposes but also for their social impact.

Continuous Sign Language Recognition (CSLR) involves detecting a sequence of glosses from the input video stream. Sign glosses can be said to be equivalent to spoken language words (with different grammar and lexicon) that match the meaning of signs \cite{cnn5}. On the other hand, Sign Language Translation (SLT) deals with generating a whole sentence from the input video stream, making it a more comprehensive and understandable version of CSLR.

Current advances in computer vision have revolutionized the way how CSLR and SLT problems are being solved. 
The ability to represent features of CNNs without hand-crafted feature extraction has made them better choices in any computer vision task. Also, the availability of large-scale datasets such as RWTH-PHOENIX-Weather \cite{rwthPhoenix2014}, extended RWTH-PHOENIX-Weather with translation corpus \cite{rwthPhoenix2014T} etc. has also contributed to the popularity of deep learning approaches for sign language problems. In some form or another, every CSLR and SLT solution includes a CNN-based feature extractor nowadays, whether it is pre-trained or trained alongside the entire model, whether it processes 2D images or 3D clips as input \cite{cnn5, cnn11, ko2019neural, min2021visual, pu2019iterative}. In CSLR, a sequence of glosses is then predicted from the features of the images or clips after some transformations. Recent solutions regarding SLT use some modified versions of encoder-decoder architecture \cite{cnn5, camgoz2020multi, ko2019neural} on the features extracted by the CNN feature extractor.

As we have already discussed, the most important aspect of devising a solution for CSLR or SLT is to understand the temporal transition of different parts of the body. Many uni-modal solutions have explored the use of 1D-CNN, RNN \cite{cnn31,pu2019iterative}, and Transformer Encoder \cite{cnn5, cnn31} to extract the temporal and global information from the sequential spatial frames of the video. Previous works have shown that such complex temporal modules provide a comprehensive understanding of temporal and global correlation. However, one of the easiest ways to improve such understanding is to include additional features such as optical flow, as it highlights movement-involved regions that are vital for sign detection.  

Several studies have already tried to incorporate multi-modal information along with RGB images. Jiang \etal \cite{jiang2021skeleton} trained separate networks for each modality and ensembled their independent predictions. As they were not trained together, one modality didn't have any information about the other and it potentially missed the joint co-relation among the modalities. Cui \etal used simple summation to combine RGB and optical flow features \cite{cnn11}. As the summation works element-wise, one frame of RGB could only get the multimodal information corresponding to that frame but not from the frames around it. Such approaches potentially fail to capture the inter-frame relationship across different modalities. Most recently, Chen \etal proposed a two-stream network for RGB and body point data \cite{chen2022two}. However, it trained two feature extractors(S3D model) simultaneously with the network, raising concerns about high computational complexity.

In our work, we try to address these shortcomings. We attempt to make the model better understand the cross-modal relationship while adding a minimal computational overhead. We use cross-attention \cite{badamdorj2021joint} between RGB and optical flow features as a plugin module, that can be added to any existing network to improve its performance. Also, we use soft distillation loss between the multi-modal features and two original features, \ie RGB and optical-flow features to properly utilize the features of both modalities.  Besides, we have used only one feature extractor to be trained in an end-to-end fashion in CSLR. In SLT, no feature extractor was used in an end-to-end manner. Though this reduces the computational complexity in several folds, it also affects the performance slightly. However, we were able to improve the performance from the baseline method, \ie SMKD \cite{cnn17} for CSLR and Stochastic Transformer \cite{voskou2021stochastic} for SLT, in both cases. Also, we were able to achieve competitive performance against existing approaches.

\section{Related Works}
\label{related_works}


\subsection{Continuous Sign Language Recognition (CSLR)}
The recent developments in Convolutional Nural Networks (CNNs) provided effective tools for modeling visual data. CSLR models using deep learning typically have three parts\cite{cnn31}. The first one is a visual feature extractor model. That can be either 2D CNN \cite{cnn17, cnn31} that deals with image level inputs or 3D CNNs \cite{pu2019iterative} that deals with clip level inputs. Afterward, comes a sequential module that helps incorporate some extent of global information into the features. Popular component for this module includes RNNs \cite{cnn31,pu2019iterative}, 1D-CNNs, or Transformers \cite{cnn5, cnn31}. And finally, an alignment module that uses HMMs \cite{cnn26} or CTC \cite{cnn17}. Some studies \cite{cnn11, cnn17} have explored iterative training as they discovered deep learning pipelines were inadequate in properly training the feature extractor.  To address the overfitting issue in end-to-end pipelines, Niu \etal \cite{cnn31} introduced the Visual Alignment Constraint (VAC) to enforce consistency between the visual and sequential modules.

Using end-to-end iterative training provides better performance by utilizing full potential of the feature extractor. But the drawback here is that most of the feature extractor is very computationally heavy compared to the rest of the network and it increases both training and evaluation time.

\subsection{Sign Language Translation}

Sign language translation (SLT) presents a more challenging and practical problem than continuous sign language recognition (CSLR). In contrast to CSLR's focus on predicting glosses, SLT must translate entire sentences, requiring consideration of phonology, grammar, morphology, vocabulary, discourse, and pragmatics. To address this challenge, Camg{\"o}z \etal \cite{Camgz2018NeuralSL} proposed the first neural SLT model, treating SLT as a neural machine translation problem. Ko \etal \cite{ko2019neural} normalized human key points and used them as input to a seq2seq model for SLT. After that, Camg{\"o}z \etal \cite{camgoz2020multi} then introduced a multi-channel transformer architecture capable of modeling the inter and intra-contextual relationships of multiple asynchronous information channels within the transformer network. They later applied transformers for sign language tasks for the first time \cite{cnn5}, proposing an end-to-end SLT and CSLR pipeline that does not require any real temporal information. Voskou \etal \cite{voskou2021stochastic} applied the local-winner-takes-all \cite{panousis2019nonparametric} approach instead of ReLU activation in the dense layers of the transformer network.



\subsection{Multi-Modal Learning}
Human communication is a multi-modal process that involves language, visual, and acoustic modalities. In the field of sign language recognition and translation, different types of additional modalities alongside the RGB video frames have been used in the past. Human body keypoints and optical flow have been the most used auxiliary modalities in sign language \cite{jiang2021skeleton, cnn11, chen2022two}. Moreover, depth information has also been shown to provide useful information in sign language detection \cite{ jiang2021skeleton}. Additionally, gaze tracking has also been used in some works as a multi-modal feature \cite{jebali2021vision}.

The method to merge the multi-modal features is also an essential aspect of the task. A diverse set of approaches has already been explored in previous tasks about the design of this merging module. Simple approaches such as summation \cite{cnn11} and concatenation \cite{chen2022two} of features have been the most used approaches in this regard. An ensemble of the prediction vector from the different streams was used by Jiang \etal \cite{jiang2021skeleton}. A graph-based approach has been used by Tang \etal \cite{tang2021graph}. 

\begin{figure*}
    \centering
    \begin{subfigure}{0.70\textwidth}
        \centering
        \includegraphics[width=\linewidth]{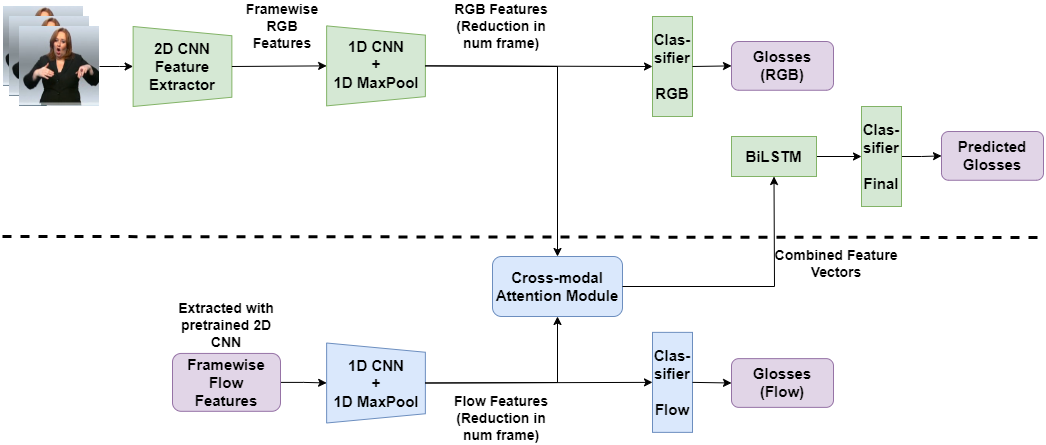}
        \caption{Cross-modal Attention with SMKD}
        \label{fig:cma_smkd}
    \end{subfigure}
    \hspace{0.3mm}
    \vline
    \hspace{0.1mm}
    \begin{subfigure}{0.28\textwidth}
        \centering
        \includegraphics[width=0.9\linewidth]{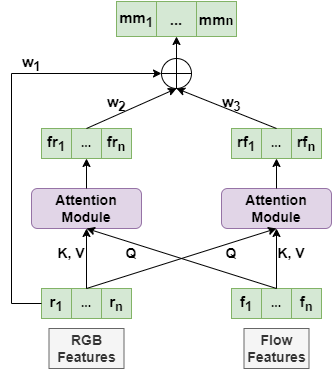}
        \caption{Cross Attention Module}
        \label{fig:attn}
    \end{subfigure}
    \caption{Cross-modal Attention on top of visual alignment constraint with self-mutual distillation learning (SMKD \cite{cnn17}) is explained in Subfigure \ref{fig:cma_smkd}. The components above the dashed line were part of the original architecture. In the original work, the Reduced RGB features were sent to the BiLSTM layer. Components below the dashed line were added to incorporate the optical flow information with RGB features. Subfigure \ref{fig:attn} illustrates the Cross Attention module.}
    \label{fig:slr_img}
\end{figure*}

\section{Methodology}
\label{method}

\subsection{Cross-Modal Attention Module}
We use a generic plug-in cross-attention extension that can be integrated into any existing network that deals with continuous sign language recognition or translation. The detailed architecture is shown in Figure \ref{fig:attn}. 

We characterize each video $V$ with a list of consecutive frames $\mathit{v}_i$, where $\mathit{i} \in 1, 2, 3, ..., n$. There are two types of data available for each frame $\mathit{v}_i$, namely RGB image, $\mathit{img}_i$, and optical flow, $\mathit{flow}_i$. We generate optical flow for each image using Flownet 2.0 \cite{ilg2017flownet}. 

We use different feature extractors for each of the modalities. These feature extractors generate frame-wise feature vectors for a specific modality. Let's denote RGB features as $\mathit{r}_i$ and flow features as $\mathit{f}_i$. We pass these two features into the cross-modal attention module. 

We follow the following steps to generate flow attended RGB features, $\mathit{fr}$. Firstly, we generate the query, $Q$, from optical flow features, $\mathit{f}$, whereas we generate key, $K$, and value, $V$, from RGB features, $\mathit{r}$.

\begin{equation}
    \begin{aligned}
    \mathit{Q} &= \mathit{W}_{f \rightarrow Q} * \mathit{f}\\
    \mathit{K} &= \mathit{W}_{r \rightarrow K} * \mathit{r}\\
    \mathit{V} &= \mathit{W}_{r \rightarrow V} * \mathit{r}
    \end{aligned}
\end{equation}
Here $\mathbf{W}$ represents the weight vector of a simple linear neural network.

The attention weights of RGB features when optical flow features are used as the query is defined as, $\mathit{\omega}_{f \rightarrow r}$

\begin{equation}
    \mathit{\omega}_{f \rightarrow r} = softmax\left(\frac{Q^TK}{\sqrt{\mathit{d}_k}}\right)
\end{equation} where $\mathit{d}_k$ is the dimension of the features.
Finally, we pass the dot product between the value, $V$, and attention wights, $\mathit{\omega}_{f \rightarrow r}$ through a dense layer to generate the final output, $\mathit{fr}$.

\begin{equation}
    \begin{aligned}
    \mathit{m} &= \mathit{V} * \mathit{\omega}_{f \rightarrow r}\\
    \mathit{fr} &= \mathit{W}_{m \rightarrow fr} * \mathit{m}
    \end{aligned}
\end{equation}

We generate the RGB features attended optical flow features, $\mathit{rf}$, in almost the same way. We just replace the query, $Q$, to be generated from RGB features, $\mathit{r}$, whereas key, $K$, value, $V$, to be generated from optical flow features, $\mathit{f}$.

Finally, we generate the multi-modal feature, $\mathit{mm}_i$, as a weighted average between the RGB features, $\mathit{r}_i$, flow attended RGB features, $\mathit{fr}_i$, and RGB features attended flow features, $\mathit{rf}_i$. The weights are made trainable with the rest of the network. 

\begin{equation}
    \mathit{mm}_i = \mathit{w}_1 * \mathit{r}_i + \mathit{w}_2 * \mathit{fr}_i + \mathit{w}_3 * \mathit{rf}_i
\end{equation}

The optical flow features, $\mathit{f}_i$, are excluded from the final multi-modal feature because it was used only to aid the representational capability of the RGB features.


\subsection{Continuous Sign Language Recognition}
For this, we use SMKD \cite{cnn17} as the baseline and add the optical flow information on top of it using the plugin explained earlier. The detailed methodology has been shown in figure \ref{fig:cma_smkd}.

In the first stage of our procedure, we train separate feature extractors, $\mathit{FE}_r$ \& $\mathit{FE}_f$, on both RGB and optical flow data respectively. The training follows a similar procedure explained in SMKD \cite{cnn17}. After the pre-training stage, we start our second stage of the procedure. 
We use both the RGB features, $\mathit{r}_i$, and optical flow features, $\mathit{f}_i$, in this stage.
\begin{equation}
    \begin{aligned}
    \mathit{r}_i &= \mathit{FE}_r (\mathit{img}_i)\\
    \mathit{f}_i &= \mathit{FE}_f (\mathit{flow}_i)
    \end{aligned}
\end{equation}
Between these feature extractors, in our second stage, we train $\mathit{FE}_r$ along with the network in an end-to-end fashion, whereas $\mathit{FE}_f$ is pre-trained using SMKD procedure and used to extract the optical flow features before the training starts.

We pass these two features into two different temporal dimension reduction networks, $\mathit{temp\_reduce}_r$ \& $\mathit{temp\_reduce}_f$. This module is made up of a combination of 1D CNN and 1D max pooling layer that reduces the number of frames in the temporal dimension approximately to a factor of $\slfrac{1}{4}$.
\begin{equation}
    \begin{aligned}
    \mathit{r\_reduced}_j = \mathit{temp\_reduce}_r (\mathit{r}_i)\\
    \mathit{f\_reduced}_j = \mathit{temp\_reduce}_f (\mathit{f}_i)
    \end{aligned}
\end{equation}
, where $\mathit{j} \in 1, 2, 3, ..., \frac{n}{4}$.

We then pass them to the cross-modal attention module to generate multi-modal features. Afterward, we use a BiLSTM layer on these multi-modal features to incorporate global context into them. Finally, we apply a classifier to each feature to generate a gloss prediction from them.

\begin{equation}
\begin{gathered}
    \mathit{mm}_j = \mathit{CrossAttention} ( \mathit{r\_reduced}_j, \mathit{f\_reduced}_j )\\
    \mathit{global}_j = \mathit{BiLSTM} ( \mathit{mm}_j )\\
    \mathit{gloss}_j = \mathit{cls}_{final} ( \mathit{global}_j ) 
\end{gathered}
\end{equation}

We also use separate classifiers after each of the $\mathit{temp\_reduce}_r$ and $\mathit{temp\_reduce}_f$ modules on different streams to generate gloss predictions from them.
\begin{equation}
    \begin{aligned}
    \mathit{gloss\_rgb}_{j} = \mathit{cls}_{rgb} ( \mathit{r\_reduced}_j )\\
    \mathit{gloss\_flow}_{j} = \mathit{cls}_{flow} ( \mathit{f\_reduced}_j )
    \end{aligned}
\end{equation}

Let's indicate the ground truth glosses as $\mathit{gt\_gloss}$. The original loss that will be used to align the predicted glossed with ground truth is Connectionist Temporal Classification(CTC) loss. Along with $\mathit{gloss}$,  $\mathit{gloss\_rgb}$ and $\mathit{gloss\_flow}$ are also supervised to be as close as the $\mathit{gt\_gloss}$.

\begin{equation} \label{eqn:ctc_losses}
    \begin{gathered}
    \mathcal{L}_{CTC} = \mathit{CTC}(\mathit{gloss}, \mathit{gt\_gloss}) \\
    \mathcal{L}_{1} = \mathit{CTC}(\mathit{gloss\_rgb}, \mathit{gt\_gloss}) \\
    \mathcal{L}_{2} = \mathit{CTC}(\mathit{gloss\_flow}, \mathit{gt\_gloss}) \\
    \end{gathered}
\end{equation}

To enable the final gloss vector to have features similar to both RGB and optical flow features, we use KL-Divergence Loss for distillation for both modalities. 
\begin{equation} \label{eqn:kd_losses}
    \begin{aligned}
    \mathcal{L}_{3} &= \mathit{KL\_Divergence}(\mathit{gloss\_rgb}, \mathit{gloss})\\
    \mathcal{L}_{4} &= \mathit{KL\_Divergence}(\mathit{gloss\_flow}, \mathit{gloss})\\
    \end{aligned}
\end{equation}
The final loss is defined as,
\begin{equation}
\label{eqn:slr_loss}
    \mathcal{L} = \mathcal{L}_{CTC} + \mathcal{L}_{1} + \mathcal{L}_{2} + \alpha \mathcal{L}_{3}  + \beta \mathcal{L}_{4},
\end{equation}where $\alpha$ and $\beta$ are already tuned for SMKD \cite{cnn17}.


\subsection{Sign Language Translation}
For this, we use Stochastic Transformer \cite{voskou2021stochastic} as the baseline and added optical flow data. The approach to adding the cross-modal attention module with sign language translation is quite simple. In this case, the model is trained with features of both RGB and optical flow modalities which have been extracted by a pre-trained network. 

\begin{figure}[h]
\begin{center}
\includegraphics[scale=0.38]{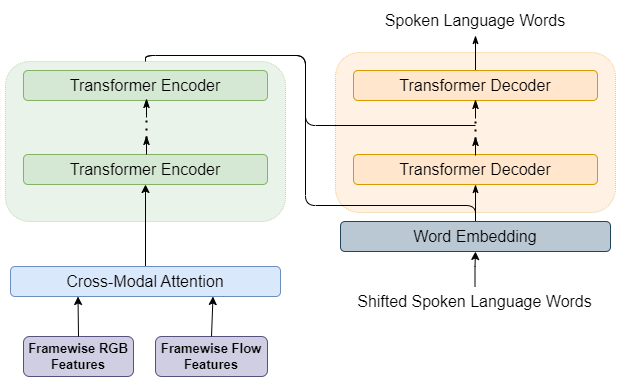}
\end{center}
   \caption{Cross-modal attention on top of stochastic transformer networks \cite{voskou2021stochastic}. Only the Cross-Modal Attention module was used additionally compared to the original pipeline.}
\label{fig:slt}
\end{figure}

We do not train any feature extractor alongside the model for this case in an end-to-end manner. RGB features and optical flow features are passed into the cross-modal attention module to generate multi-modal features. These multi-modal features are then passed onto the original network instead of the RGB features. An overview of the method has been shown in figure \ref{fig:slt}.

\begin{equation}
    \begin{aligned}
    \mathit{r}_i &= \mathit{FE}_r (\mathit{img}_i)\\
    \mathit{f}_i &= \mathit{FE}_f (\mathit{flow}_i)
    \end{aligned}
\end{equation}
\begin{equation}
    \mathit{mm}_i = \mathit{CrossAttention} ( \mathit{r}_i, \mathit{f}_i )
\end{equation} 
\begin{equation}
    \mathit{pred\_tokens} = \mathit{Encode\_Decoder\_with\_LWTA} ( \mathit{mm})
\end{equation}

In this scenario, we train the model with cross-entropy loss between predicted tokens and ground truth tokens.
\begin{equation}
    \mathcal{L} = \mathit{CrossEntropy} ( \mathit{predicted\_tokens}, \mathit{gt\_token} )
\end{equation}


\section{Experiment Results}
\subsection{Continuous Sign Language Recognition} 
We use SMKD \cite{cnn17}, which originally used only RGB modality, as a baseline for the experiments of CSLR. In the final version, on top of the baseline, we have only added the cross-attention module and a temporal dimension reduction module, which is just a combination of 1D CNN and MaxPool operation. 

For evaluation of our approaches, we have used the RWTH-PHOENIX-Weather \cite{rwthPhoenix2014} dataset to report the results of the CSLR experiments. Along with the video, this dataset provides the sequence of ground truth glosses for a specific video. This dataset has a total of 5671 samples for training, 540 samples for evaluation, and 629 samples for testing. We have used the WER metric \cite{cnn11} for evaluation.

\subsubsection{Ablation Study}
\textbf{Experiment with different modality:}
In this experiment, we consider two modalities of data, original RGB images, and corresponding optical flow images. We experiment with all the combinations of these two types of features to understand their effect. We have used the SMKD \cite{cnn17} approach on only RGB, only optical flow, optical flow attended RGB, and RGB attended optical flow. Their results have been listed in table \ref{tab:modality} 
\begin{table}[h]
\begin{center}
\begin{tabular}{|c|c|c|}
\hline
\multirow{2}{*}{Data Modality} & \multicolumn{2}{c|}{WER}\\
\cline{2-3}

& Dev & Test \\
\hline
RGB & 19.8 & 20.5 \\
\hline
Optical Flow & 25.5 & 27.2 \\
\hline
RGB attended by Optical Flow & 19.7 & 20.2 \\
\hline
Optical Flow attended by RGB & 23.9 & 25.1 \\
\hline
\end{tabular}
\end{center} 
\vspace{-12pt}
\caption{Effect on WER with different data modality SMKD}
\label{tab:modality}
\end{table}

Evidently, RGB attended by optical flow outperforms RGB features, and optical flow attended by RGB outperforms optical flow. It is clear from the experiment that multi-modal features perform better than their single-modal counterparts.

\textbf{Experiment the feature extractor:} 
Feature extraction is the most crucial part of this entire solution. For this reason, we have experimented with different feature extractors. The results have been shown in Table \ref{tab:feature_extract}.

\begin{table}[h]
\begin{center}
\begin{tabular}{|c|c|c|c|}
\hline
\multirow{2}{*}{Feature Extractor} & \multirow{2}{*}{Feature Size} & \multicolumn{2}{c|}{WER}\\
\cline{3-4}

& & Dev & Test \\
\hline
ResNet18 & 512 & 19.8 & 20.5 \\
\hline
EfficientNet-B0 & 1280 & 21.3 & 23.1 \\
\hline
Inception-v4 & 1536 & 23.4 & 25.8 \\
\hline
ResNet34 & 512 & 19.6 & 20.4 \\
\hline
\end{tabular}
\end{center} 
\vspace{-12pt}
\caption{Effect on WER with varying CNN-based feature extractors on SMKD}
\label{tab:feature_extract}
\end{table}

It can be seen that, as the extracted features get more complex the network starts to overfit.

\textbf{Comparison with other multi-modal approaches:}
We have already discussed the approaches adopted previously to combine multi-modal features in Section \ref{related_works}. Ensembling of predictions, Summation, and concatenation of features have been the most used merging techniques in previous works. To prove cross-attention is a better feature merger than the already applied approaches in CSLR, we have compared their performance in table \ref{tab:multi-mod}.

\begin{table}[h]
\begin{center}
\begin{tabular}{|c|c|c|}
\hline
\multirow{2}{*}{Different Cross-Modal Approaches} & \multicolumn{2}{c|}{WER}\\
\cline{2-3}

& Dev & Test \\
\hline
Ensembling & 22.6 & 23.2 \\
\hline
Summation & 20.0 & 21.3 \\
\hline
Concatenation & 20.5 & 20.9 \\
\hline
Cross-Modal Attention & 19.6 & 19.6 \\
\hline
\end{tabular}
\end{center} 
\vspace{-12pt}
\caption{Comparison of different procedures of combining multi-modal features on SMKD.}
\label{tab:multi-mod}
\end{table}

\textbf{Experiment on the stage when to apply the cross-modal attention: }
There are two possibilities regarding when to apply the cross-modal attention module. The first one is right after the 2D CNN feature extraction and another is after temporal reduction (1D CNN and 1D max pooling operation). We have used the exact same components as shown in Figure \ref{fig:cma_smkd} for this experiment. The comparison between them is shown in Table \ref{tab:stage}. The result points out that applying the cross-attention on temporally consolidated features provides better performance. This also points out the necessity of a temporal reduction module to extract the important frames.

\begin{table}[h]
\begin{center}
\begin{tabular}{|c|c|c|}
\hline
\multirow{2}{*}{Stage to apply Cross-modal Attention} & \multicolumn{2}{c|}{WER}\\
\cline{2-3}

& Dev & Test \\
\hline
After 2D CNN Feature Extractor & 20.9 & 21.4 \\
\hline
After 1D CNN and Max Pooling & 19.7 & 20.1 \\
\hline
\end{tabular}
\end{center} 
\vspace{-12pt}
\caption{Effect on WER based on the stage after which cross-modal attention is applied. Pretrained ResNet18 is used as a feature extractor on both modalities.}
\label{tab:stage}
\end{table}
\vspace{-10pt}
\textbf{Comparison with existing Methods:} 
In Table \ref{tab:slrdat}, we have compared existing CSLR approaches against our cross-attention approach and it shows our approach achieved better test scores. Our approach reduced the relative WER by 4.3\% from the baseline, SMKD \cite{cnn17}. Apart from performing better than the baseline, we have also compared the performance against some recent methods. 

\begin{table*}
\begin{center}
\begin{tabular}{|c|c|c|c|c|c|c|c|c|}
\hline
Depth & \multicolumn{4}{c|}{Dev} & \multicolumn{4}{c|}{Test}\\
\cline{2-9}

Cross - Enc - Dec& BLEU-1 & BLEU-2 & BLEU-3 & BLEU-4 & BLEU-1 & BLEU-2 & BLEU-3 & BLEU-4 \\
\hline

1-1-1 & 49.0 & 35.12 & 27.38 & 21.96 & 46.92 & 35.08 & 27.24 & 22.20 \\
\hline
\textbf{1-2-2} & \textbf{49.52} & \textbf{36.08} & \textbf{28.16} & \textbf{23.04} & \textbf{48.62} & \textbf{36.16} & \textbf{28.45} & \textbf{23.42} \\
\hline
1-3-3 & 46.11 & 33.66 & 26.23 & 22.01 & 46.29 & 34.07 & 26.39 & 22.10 \\
\hline
2-2-2 & 47.31 & 34.58 & 27.34 & 22.12 & 47.16 & 34.79 & 27.09 & 21.98 \\
\hline
2-3-3 & 45.92 & 33.38 & 25.68 & 21.74 & 45.80 & 33.67 & 24.91 & 21.55 \\
\hline
\end{tabular}
\end{center} 
\vspace{-12pt}
\caption{Comparison between different depths of cross-modal attention, encoder and decoder.}
\label{tab:depth}
\end{table*}


\begin{table*}
\begin{center}
\begin{tabular}{|c|c|c|c|c|c|c|c|c|}
\hline
\multirow{2}{*}{Method} & \multicolumn{4}{c|}{Dev} & \multicolumn{4}{c|}{Test}\\
\cline{2-9}

 & BLEU-1 & BLEU-2 & BLEU-3 & BLEU-4 & BLEU-1 & BLEU-2 & BLEU-3 & BLEU-4 \\
\hline
Sign2Gloss2Text \cite{Camgz2018NeuralSL} & 44.40 & 31.83 & 24.61 & 20.16 & 44.13 & 31.47 & 23.89 & 19.26\\
\hline
BERT2BERT \cite{de2021frozen} & - & - & - & 21.26 & - & - & - & 21.16\\
\hline
BERT2RND \cite{de2021frozen} & - & - & - & 22.47 & - & - & - & 22.25\\
\hline
Stochastic Transformer \cite{voskou2021stochastic}   & 49.66 & 35.65 & 27.74 & 22.55 & 48.54 & 35.50 & 27.65 & 22.63 \\
\hline
\textbf{Ours} & 49.52 & \textbf{36.08} & \textbf{28.16} & \textbf{23.04} & \textbf{48.62} & \textbf{36.16} & \textbf{28.45} & \textbf{23.42} \\
\hline
\end{tabular}
\end{center} 
\vspace{-12pt}
\caption{Comparison between stochastic transformer network with cross-modal attention approach and some existing approaches. For our method, the best combination of layer depths was used.}
\label{tab:sltdat}
\end{table*}


\begin{table}[h]
\begin{center}
\begin{tabular}{|c|c|c|}
\hline
\multirow{2}{*}{Method} & \multicolumn{2}{c|}{WER}\\
\cline{2-3}

& Dev & Test \\
\hline
Stochastic CSLR \cite{cnn31} & 24.9 & 25.3 \\
\hline
Cross-modal Alignment \cite{papastratis2020continuous} & 23.9 & 24.0 \\
\hline
VAC \cite{min2021visual} & 21.2 & 22.3 \\
\hline
STMC \cite{zhou2020spatial} & 21.1 & 20.7 \\
\hline
SMKD \cite{cnn17} & 19.8 & 20.5 \\
\hline
SKMD with ResNet34 & 19.6 & 20.4 \\
\hline
C2SLR \cite{zuo2022c2slr} & 20.2 & 20.4\\
\hline
SignBERT+ \cite{hu2023signbert} & 19.9 & 20.0 \\
\hline
RadicalCTC \cite{min2022deep} & 19.4 & 20.2\\
\hline
\textbf{Ours} & 19.6 & \textbf{19.6} \\
\hline
\end{tabular}
\end{center} 
\vspace{-12pt}
\caption{Performance comparison between the existing CSLR approach with our proposed method. On top of SMKD \cite{cnn17} with ResNet34, we add the cross-modal attention module for feature merging.}
\label{tab:slrdat}
\end{table}


\vspace{-5pt}
\subsection{Sign Language Translation} 
We use Stochastic-Transformer \cite{voskou2021stochastic}, which originally used only RGB modality, as a baseline for the experiments of SLT. In the final version, on top of the baseline, we have only added the cross-attention module. For evaluation of our approaches, we have used the RWTH-PHOENIX-Weather with translation corpus \cite{rwthPhoenix2014T} dataset to report the results of the SLT experiments. Along with the video, this dataset provides a complete sentence as a label for the video. This dataset has a total of 7096 samples for training, 519 samples for evaluation, and 642 samples for testing. We have used BLEU Score \cite{papineni2002bleu} for evaluating SLT performance.

\subsubsection{Ablation Study}
\textbf{Depth of the encoders and decoder:}
In transformer networks, generally, multiple encoders and decoders are stacked on top of each other. How many of them should be stacked together is solely case-dependent. Our proposed cross-attention module can also be stacked one after another. For this reason, it is necessary to figure out the best combination of these components. Effect of varying depth has been shown in table \ref{tab:depth}.

\textbf{Comparison with existing Methods:} 
In Table \ref{tab:sltdat}, we have compared a few existing SLT approaches against our cross-attention approach and it shows our approach achieved better test scores than them. As our approach doesn't include any feature extractor in an end-to-end training scheme, we have only compared against the approaches of such kind. Our approach improved the relative BLEU-4 by 3.5\% from the baseline, Stochastic-Transformer \cite{voskou2021stochastic}.

\section{Conclusion}
In this work, we use cross-attention between RGB features and optical flow features to generate a unified multi-modal feature representation that contains the movement-related information from optical flow along with usual RGB features. The additional plugin cross-modal module is very lightweight (only two attention modules). Besides, our two-stage training allows us to use only one feature extractor in CSLR and no feature extractor in SLT for end-to-end training and inference. We have used the same module on two different works establishing that it can be integrated into any uni-modal procedure. Furthermore, in both cases, we witnessed improvement in performance on the benchmark dataset. In future work, we can also explore the effect of other modalities such as body key points or depth information. We can also try other types of attention mechanisms for feature mergers.

\section{Acknowledgements}
This research was partially supported by Bangladesh University of Engineering and Technology (BUET), Dhaka, Bangladesh and ICT Division, Government of People's Republic of Bangladesh.

\small
\bibliographystyle{IEEEbib}
\bibliography{refs}

\end{document}